# Applying Machine Learning for Duplicate Detection, Throttling and Prioritization of Equipment Commissioning Audits at Fulfillment Network

**Farouq Halawa, Majid Abdul, Raashid Mohammed**
**Global Engineering Support Services at Amazon.com**

## Abstract

VQ (Vendor Qualification) and IOQ (Installation and Operation Qualification) audits are implemented in warehouses to ensure all equipment being turned over in the fulfillment network meets the quality standards. Audit checks are likely to be skipped if there are many checks to be performed in a short time. In addition, exploratory data analysis reveals several instances of similar checks being performed on the same assets and thus, duplicating the effort. In this work, Natural Language Processing and Machine Learning are applied to trim a large checklist dataset for a network of warehouses by identifying similarities and duplicates, and predict the non-critical ones with a high passing rate. The study proposes ML classifiers to identify checks which have a high passing probability of IOQ and VQ and assign priorities to checks to be prioritized when the time is not available to perform all checks. This research proposes using NLP-based BlazingText classifier to throttle the checklists with a high passing rate, which can reduce 10%-37% of the checks and achieve significant cost reduction. The applied algorithm over performs Random Forest and Neural Network classifiers and achieves an area under the curve of 90%. Because of imbalanced data, down-sampling and up-weighting have shown a positive impact on the models' accuracy using F1 score, which improved from 8% to 75%. In addition, the proposed duplicate detection process identifies 17% possible redundant checks to be trimmed.

## 1. Introduction

The use of machine learning (ML) is becoming a norm in warehousing solutions. Recent developments related to maintenance engineering enabled a broad range of ML applications spanning from storage allocation, material handling systems [1,2] and asset management [3]. Predictive maintenance aims at applying machine learning and probabilistic modeling to estimate the time when a maintenance check should be implemented, which improves safety and avoids costly repairs [4]. This research looks into a new application in predictive maintenance, which includes the prioritization, throttling, and duplicate detection of auditing checks in warehousing networks using a case study. The research study targets the process used for validating the quality and functionality of material handling equipment and base building management assets in warehouses before releasing them to operations. Two types of maintenance audits are typically performed for the assets in a warehouse; one by the vendor as part of a Vendor Qualification (VQ) process. The process insures that the assets that exist in a newly constructed warehouse confirm the vendor standards. The second auditing process is called Installation and Operation Qualification (IOQ), which is conducted regularly by the maintenance teams. A recent US market study [5] shows that current labor shortages within the US have driven down the number of experienced skilled workers available for hire to complete mechanical inspection work. Meanwhile, there is an increase in the number of constructed warehouses to align with the concept of just-in-time, which causes an increase in the number of audit checks required in a warehousing network under limited human workforce. Therefore, optimizing the number of audit checks is a crucial step to avoid VQ false-positives (i.e., a vendor passes a check but a more experienced maintenance team fails the same check) and skipping rates of checks (i.e., some audit checks are bypassed due to non-stuffiest time). Identifying redundant checks and eliminating or consolidating them can cause improvement to data quality, reduction of duplication efforts, and better implementation of ML techniques. Throttling of concise checklist will also speed up the commissioning process, thereby, reduce labor costs.

This research proposes a dynamic approach to facilitate the review process of assets in warehousing networks, by predicting the checks that will pass or fail prior to implementing the checks, and thus streamlining the review process. Unlike recent studies in predictive maintenance [1,3] which focused on predicting the time needed for implementing the checks, this study looks into applying ML for reducing the volume of the maintenance checks using machine learning. The main contribution of this work is that it provides a novel application of ML in the field of maintenance using a case study, by focusing on the duplication of effort and the reduction of audit checks. This study applies non-supervised learning to cluster text data for duplicate detection and throttling, and supervised learning for predicting checklist audits that have high passing rates.

## 2. Methodology

This section outlines the proposed research framework.

2.1. Framework for throttling and prioritization

The proposed framework (described in Figure 1) was scripted on Python 3.0 to predict the checklists that can be throttled (trimmed) for future fulfillment centers. To come up with the proposed research framework, a sampled 4M checklist dataset was used. The main features are summarized in Table 1.

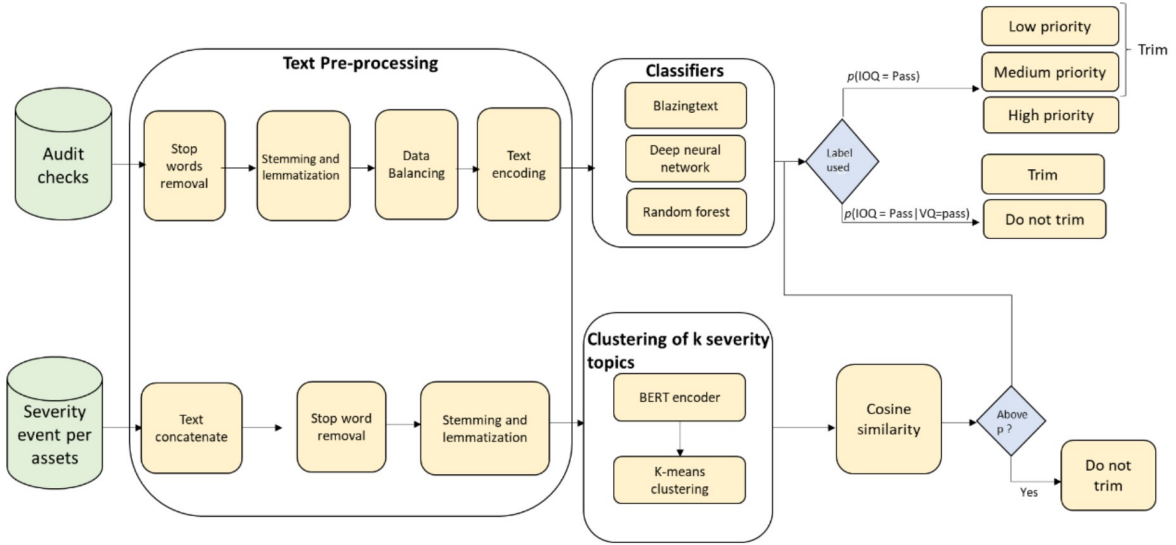

**Figure 1:** NLP-based algorithm for throttling and prioritization of audit checks

The first step of the algorithm is to pre-process the text data by removing missing values. For each word of the checklist sentences stemming and lemmatization were implemented. The next step was to vectorize the text data, which was different based on the used classifier. In this research, three classifiers are tested (explained in Section 2.1.1).

*2.1.1 Tested classifiers*: The first algorithm is BlazingText [6], which can provide highly optimized implementations of the Word2vec and text classification. Word2vec represents each word $w$ in a vocabulary $V$ of size $T$ as a low-dimensional dense vector $v_w$. To put this formally, given a large training corpus represented as sequence of words $w_1, w_2, \ldots w_N$ the objective is to maximize the log-likelihood, where T is the vocabulary size and the context $c$ is set of indices of words containing a word $w_c$ (Equation 1). Once the data is vectorized, it is fed into a shallow neural network to predict the classification labels. The classifier is explained in Figure 2(a).

$$\text{log likelihood} = \sum_{t=1}^{T} \sum_{c=1}^{N} \log p(w_c | w_t) \qquad (1)$$

BlazingText utilizes Word2vec encoder, which does not take into the account the word position or order. An alternative approach is to use BERT encoder [7], which stands for Bidirectional Encoder Representations from Transformers. BERT is designed to pre-train deep bidirectional representations from unlabeled texts by jointly conditioning on both left and right contexts in all layers, thus, it considers the syntax of the sentences. The proposed deep neural network uses BERT encoding in addition to count encoder for the categorical variables and concatenates that into 3 dense layers, as shown in Figure 2(b). In both classifiers a and b, Softmax function (Equation 2) is used to calculate the probability distributions. The Softmax function takes as input a vector $z$ of $K$ real numbers, and normalizes it into a probability distribution consisting of K probabilities proportional to the exponentials of the input numbers. Keras Tensorflow was used to code classifier b, while SageMaker was used in classifier a. The final classifier (Figure 2(c)) uses a different approach for encoding, by constructing a large matrix of the top words in each feature and then calculate the count for each word. Then, Random Forest [8] is used, which utilizes the concept of bagging and decision trees. The algorithm fits a number of decision tree classifiers on various sub-samples of the dataset. To train the machine learning model, the data was split into training and testing sets of 80% and 20%, respectively. Evaluation metrics are explained in Table 2. The proposed algorithms can be used for two tasks: predicting passing IOQ given VQ is passed, or assigning priorities to the checklists using their classification probability. This research will compare between these two approaches.

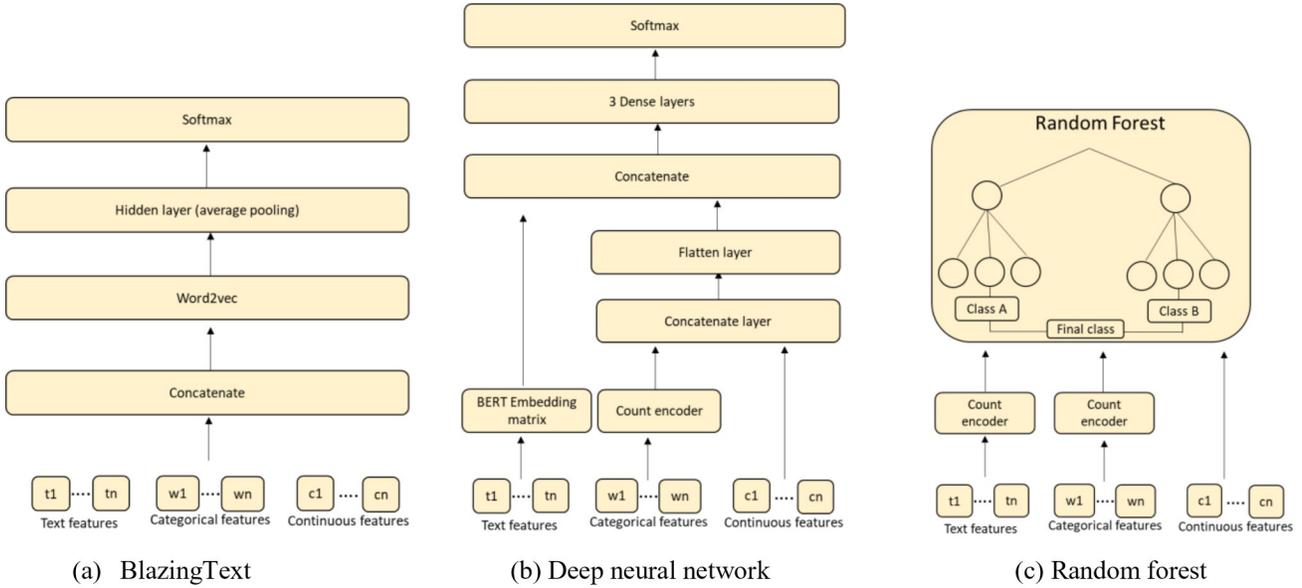

(a) BlazingText  (b) Deep neural network  (c) Random forest
**Figure 2:** Structure of tested classifiers

$$\sigma(z_i) = \frac{e^{z_i}}{\sum_i^K e^{z_i}} \quad (2)$$

**Table 1**: Features and labels of the dataset

| Features | Name | Description | Data type |
|---|---|---|---|
| 1 | Asset type | The type of asset on which the audit was performed | Categorical |
| 2 | Vendor | The vendor of the asset | Categorical |
| 3 | Checklist | The audit check that is performed | Text |
| 4 | Focus points | The check may be applied to different focus point per asset | Text |
| 5 | Criticality | The criticality of the checklist | Categorical |
| 6 | Severity score | The likelihood that a checklist causes a severity event | Continuous |
| 7 | Severity group | The mapped severity group per check | Categorical |
| **Labels** | **Name** | **Description** | **Data type** |
| 1 | Throttled | If a check passes both IOQ and VQ audits then it can be throttled | Binary |
| 2 | IOQ status | The status of the IOQ check include a pass or fail | Binary |

*2.1.2 Severity scoring:* To avoid the case that throttled checks may cause severity events, a non-supervised learning algorithm is proposed to estimate the likelihood of creating a severity event if an audit check is trimmed. This is implemented throughout the second stage of the algorithm "clustering" shown in Figure 1. The main idea is to use BERT encoder to vectorize the severity event dataset, then due to a large data size, cluster the severity event into groups using k-means. Elbow diagram is used to select the optimal number of clusters. Each cluster contains a group of severity events. Then, cosign similarity is applied to quantify the level of similarity between the checklist and the severity group clusters using Equation (3). The main approach to detect a similarity between two pairs is to calculate the angle between the two BERT vectors A and B, such that each vector contains the weights of the BERT hidden layer (i.e., A = $[w_1, w_1,..]$). If the angle value, $\theta$ is 0, the cosine value is 1, and thus, they are parallel to each other (i.e., similar). If they are perpendicular to each other, then the angle = 90 degrees and the cosign similarity is 0 (Figure 3). In other words, if there is a severity risk, then throttling does not occur. Otherwise, the checklist is trimmed.

$$Cos(\theta) = \frac{A^T.B}{|A||B|} \quad (3)$$

**Table 2**: Used accuracy metrics

| Metrics | Equations |
|---|---|
| 1 | $Accuracy = \frac{True\ positive + True\ negative}{Positive + Negative}$ |
| 2 | $Sensitivity = \frac{True\ positive}{True\ positive + False\ negative}$ |

| 3 | $Specificity = \dfrac{True\ negative}{True\ negative\ +\ False\ positive}$ |
| 4 | $Area\ under\ the\ curve\ (AUC) = \int_{x1}^{xn} f(x)dx$ |
| 5 | $F1\ score = \dfrac{True\ positive}{True\ positive + 0.5(False\ positive + False\ negative)}$ |

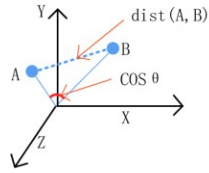

**Figure 3**: Cosine similarity concept for checklist A and B

2.2. Duplicate detection using topic modeling and cosine similarity

The following section outlines the methodology to identify and remove duplicates. For instance, an example of two duplicate audit checks are "All breakers in panel are visually damage free" and "Breaker is visually free of damage". To pre-process the text data, an algorithm was coded on python 3.0 to remove stop words, fill out missing data, stem and lower case the text data, and remove possible typos. The information about checklist, vendor, teams, focus point, and asset types are then converted into vectors using count-vectorizer. Fuzzy match algorithm was then implemented on each pair of checklists using cosine similarity (Equation 3). Similarity is defined if two checks have the same asset type, same vendor, and same fulfillment center (Features explained in Table 1). Subsequent to that, the thresholds for similarity were defined as 35% for checklists to be similar using 3 levels; if the number is between 85%-100%, then the two checks are duplicates, if the number is between 60%-85% then the two checks are highly similar, and if the level is between 35%-59% then the similarity is moderate. Those values were estimated based on experimentations.

## 3. Experimental Results

3.1. Comparison of classification algorithms

The accuracy results of the three tested ML classifiers, described in Figure 1, are summarized in Table 3. The results show that BlazingText has the highest area under the curve (AUC) of 90.2%. The BlazingText algorithm was also higher in terms of its sensitivity. The model accuracy was 83.9%. Figure 4 visualizes AUC for the 3 tested algorithms. In addition, it shows the impact of the different text features on AUC, as explained in Table 1. The feature selection process shows that the model's accuracy improved when including vendor name, asset number, focus points, and criticality texts. Table 4 shows the optimal balancing ratio of the two binary classes, using the label as passing IOQ/VQ checks. The concept of Up-weighting was also used by giving a higher weight to the minority class. F1 score was used to select the best values of the two classes (i.e., number of fail class and number of pass class). The improved F1 score is due to a better capability of the algorithm to predict the class associated with failing the checks.

**Table 3**: Accuracy scores for the tested text classifiers

| Tested Algorithms | Metrics | | | | |
| --- | --- | --- | --- | --- | --- |
|  | AUC | Accuracy | Sensitivity | Specificity | Training time (min) |
| BlazingText | 90.20% | 83.90% | 71.20% | 96.50% | 5.70 min |
| Deep neural network | 89.00% | 83.28% | 69.06% | 97.40% | 27.2 min |
| Random Forest | 84.70% | 79.70% | 62.68% | 96.70% | 4.00 min |

**Table 4**: Optimal data balancing ratio using Random Forest classifier

| Experiment | Up-weighting | # Fail class | # Pass class | Balancing ratio | F1 score (fail class) |
| --- | --- | --- | --- | --- | --- |
| 1 | No | 63,212 | 1,000,000 | 15/1 | 8% |
| 2 | Yes | 63,212 | 1,000,000 | 15/1 | 25% |
| 3 | No | 63,212 | 100,000 | 1.5/1 | 65% |
| 4 | Yes | 63,212 | 100,000 | 1.5/1 | 70% |
| 5 | No | 63,212 | 100,000 | 1/1 | 74% |
| 6 | Yes | 63,212 | 100,000 | 1/1 | 75% |

The classification algorithm was also used to prioritize the importance of checklists, as shown in Table 5 using the prediction probability of the classifier. The threshold of the predicted probability was used to select the three priority levels.

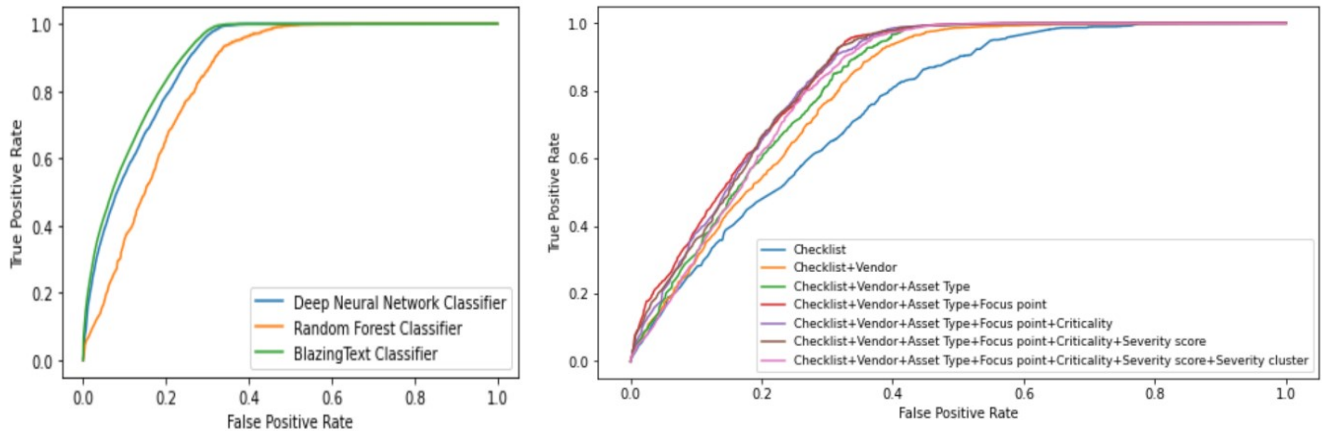

(a) Comparison between three classifiers  (b) AUC for different text features using random forest

**Figure 4:** Area under the curve for tested throttling classifiers

Table 5: Proposed levels of prioritization

| Priority Level | Threshold | Sensitivity | Specificity |
|---|---|---|---|
| Level 1: Low Priority | >90% | 97% | 35% |
| Level 2: Medium Priority | 79%-90% | 90% | 55% |
| Level 3: High priority | <72% | 86% | 64% |

3.2. Prediction of severity events

Figure 5(a) summarizes the k-means clustering of the used severity events. The elbow diagram suggests to use 14 clusters. The cosine similarly scores for mapping checklists and severity events are shown in Figure 5(b). After the clusters are obtained, cosign similarity scores are calculated. The distribution in Figure 5(b) shows that some checks are highly associated to severity events, while others are not. Using this approach, new audit checks are compared against the clusters and the similarity score is calculated for each.

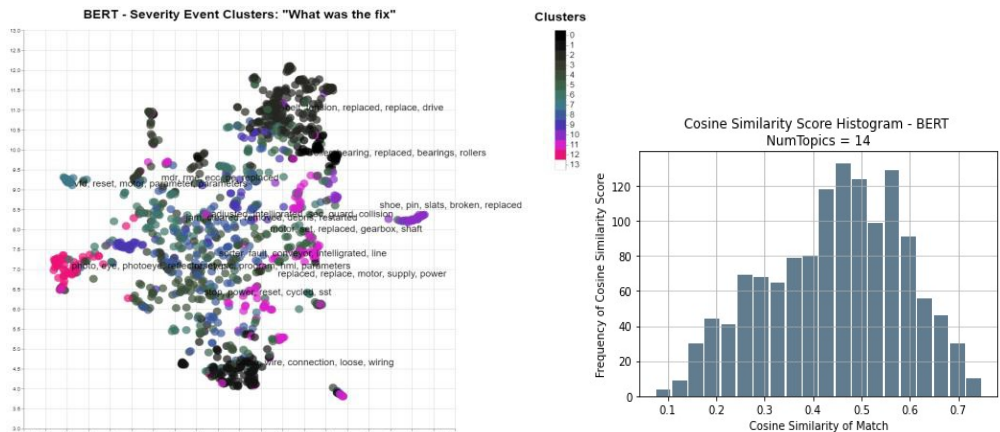

(a) Severity event clusters  (b) cosign similarity distribution

**Figure 5:** Clustering results of severity event data

3.3. Results of throttling, prioritization, and duplicate detection

The results of running the framework (Figure 1) are summarized in Figure 6, and show that the use of a similarity score, t < 0.50 results in 10% of checks being trimmed using passing IOQ/VQ as a classifier's label. The prioritization algorithm can trim between 23%-37% of the warehouse network dataset. The possibility to trim checks declines significantly if a lower similarity threshold $t$ is used (e.g., $t < 0.30$). Results are summarized in Figure 6. This reduction in checks can translate into cost saving, since less workforce will be needed to implement the non-trimmed checks. Furthermore, the duplicate detection methodology applied on the test dataset results in a 17% extra reduction due to similar/duplicates checks, as shown in Table 6.

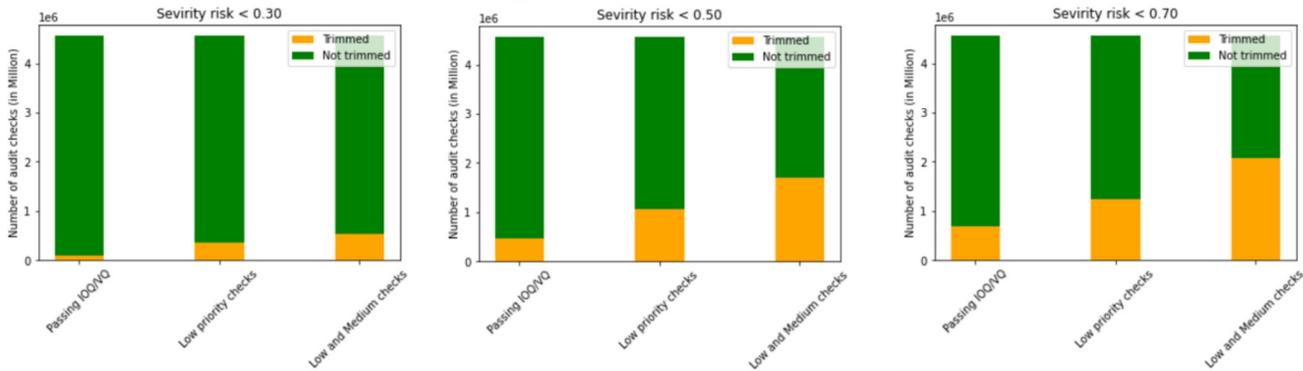

**Figure 6**: Throttling results for various servility scores and classification labels

Table 6: Identified duplicates and similar checks

| Category | Total identified checks | |
| --- | --- | --- |
| | Number | Percentage |
| Identical | 254,164 | 6% |
| High similarity | 170,903 | 3% |
| Medium similarity | 324,278 | 7% |
| **Total** | **749,345** | **17%** |

## 4. Conclusion and Future Research

In this work, ML was used to guide and facilitate the review process of the maintenance checklists that are used by a large fulfillment network. The study provided a non-supervised learning approach to identify checklists with a high similarity. Using historical data, it was possible to identify 17% of checks as being duplicates or sharing high similarity. In addition, supervised learning shows that up to 37% of checks could have been throttled in 2010-2021 since these tests had a high passing rate. With new emerging fulfillment centers and new checks being constantly added, the proposed NLP-based ML has been deployed to an app to facilitate the trimming process and to control and monitor the auditing process. This work shows that BlazingText classifier is an efficient classifier which overperforms the tested deep neural network and random forests. One limitation of the existing training data is that it contains many duplicate or similar checks. Furthermore, the size of the sentences is limited to 20 words each, which compromises the capabilities of BERT and word2vec encoders. The study shows a minor difference between the two methods using the tested dataset. Future work will implement the proposed methodology on other type of datasets which tend to have longer text sentences.

### Acknowledgments

The authors of this work would like to thank Jen Amlani for her contribution in the severity scoring method and Figure 5. In addition, thanks to Kristine Vacola from Amazon for her valuable insights and data acquisition.